\documentstyle[11pt,acl,fullname]{article}
\newcommand{\bit}{\begin{itemize}}
\newcommand{\eit}{\end{itemize}} \newcommand{\ben}{\begin{enumerate}}
\newcommand{\een}{\end{enumerate}} \newcommand{\bqt}{\begin{quote}}
\newcommand{\eqt}{\end{quote}} \newcommand{\bc}{\begin{center}}
\newcommand{\ec}{\end{center}} \newcommand{\bdes}{\begin{description}}
\newcommand{\edes}{\end{description}}
\newcommand{\btab}{\begin{tabular}} \newcommand{\etab}{\end{tabular}}

\newcommand{\ra}{$\rightarrow$}

\title{\vspace{-0.5in}
Recognizing Referential Links:\\ An Information Extraction Perspective}
\author{Megumi Kameyama\\Artificial Intelligence Center\\
SRI International\\
333 Ravenswood Ave., Menlo Park, CA 94025, U.S.A.
\\{\tt megumi@ai.sri.com}}

\begin{document}
\bibliographystyle{fullname}
\maketitle
\input{psfig}
\vspace{-0.5in}
\begin{abstract}
We present an efficient and robust reference resolution algorithm in
an end-to-end state-of-the-art information extraction system, which
must work with a considerably impoverished syntactic analysis of the
input sentences. Considering this disadvantage, the basic setup to
{\em collect, filter}, then {\em order by salience} does remarkably
well with third-person pronouns, but needs more semantic and discourse
information to improve the treatments of other expression types.
\end{abstract}

\section{Introduction}

Anaphora resolution is a {\em component technology} of an overall
discourse understanding system.  This paper focuses on reference
resolution in an information extraction system, which performs a
partial and selective `understanding' of unrestricted discourses.

\section{Reference Resolution in IE}

An information extraction (IE) system automatically extracts certain
predefined target information from real-world online texts or speech
transcripts.  The target information, typically of the form ``who did
what to whom where when,'' is extracted from natural language
sentences or formatted tables, and fills parts of predefined template
data structures with slot values.  Partially filled template data
objects about the same entities, entity relationships, and events are
then merged to create a network of related data objects.  These
template data objects depicting instances of the target information
are the raw output of IE, ready for a wide range of applications such
as database updating and summary generation.\footnote{The IE
technology has undergone a rapid development in the 1990s driven by the
series of Message Understanding Conferences (MUCs) in the
U.S. government-sponsored TIPSTER program ({\tt
http://www.tipster.org}). }

In this IE context, reference resolution takes the form of {\em
merging} partial data objects about the same entities, entity
relationships, and events described at different discourse positions.
Merging in IE is very difficult, accounting for a significant portion
of IE errors in the final output.  This paper focuses on the
referential relationships among {\em entities} rather than the more
complex problem of {\em event} merging.

An IE system recognizes particular target information instances,
ignoring anything deemed irrelevant.  Reference resolution within IE,
however, cannot operate only on those parts that describe target
information because anaphoric expressions within target
linguistic patterns may have antecedents outside of the
target, and those that occur in an apparently irrelevant pattern may
actually resolve to target entities. For this reason, reference
resolution in the IE system needs  access to {\em all} of the
text rather than some selective parts. Furthermore, it is
reasonable to assume that a largely {\em domain-independent} method of
reference resolution can be developed, which need not be tailored anew
each time a new target is defined.

In this paper, I discuss one such entity reference resolution
algorithm for a general geo-political business domain developed for
SRI's FASTUS$^{TM}$ system \cite{fastus}, one of the leading IE
systems, which can also be seen as a representative of today's IE
technology.

\subsection{The Input to Reference Resolution}

Multiple top-scoring sites working on IE have converged on the use of
finite-state linguistic patterns applied in stages of smaller to
larger units.  This finite-state transduction approach to IE, first
introduced in SRI's FASTUS, has proven effective
for real-world texts because full parsing is far too ambiguous, slow,
and brittle against real-world sentences.  This means that we cannot
assume correct and full syntactic structures in the input to reference
resolution in a typical IE system. The input is a set of (often
overlapping or discontiguous) finite-state approximations of sentence
parts.  We must {\em approximate} fine-grained theoretical proposals
about referential dependencies, and adapt them to the context of
sparse and incomplete syntactic input.

The input to reference resolution in the theoretical literature is
assumed to be fully parsed sentences, often with syntactic attributes
such as grammatical functions and thematic roles on the constituents
\cite{Webber:PhD,Sidner:Towards,hobbs:1978,GJW:1995}.  
In implemented reference
resolution systems, for pronoun resolution in particular, there seems
to be a trade-off between the completeness of syntactic input and the
robustness with real-world sentences. In short,
more robust and partial 
parsing gives us wider coverage, but less syntactic
information also leads to less accurate reference resolution.  For
instance, \namecite{lappin+leass:1994} report an 86\% accuracy for a
resolution algorithm for third-person pronouns using fully parsed
sentences as input. \namecite{kennedy+boguraev:1996} then report a
75\% accuracy for an algorithm that approximates Lappin and Leass's
with more robust and coarse-grained syntactic input. After describing
the algorithm in the next section, I will briefly compare the present
approach with these pronoun resolution approaches.

\section{Algorithm}

This algorithm was first implemented for the MUC-6 FASTUS system
\cite{fastus4}, and produced one of the top scores (a recall of 59\%
and precision of 72\%) in the MUC-6 Coreference Task, which evaluated
systems' ability to recognize coreference among noun phrases
\cite{beth-muc6}.  Note that only {\em identity} of reference was
evaluated there.\footnote{Other referential relationships such as subset
and part-whole did not reach sufficiently reliable interannotator
agreements. 
Only identity of reference had a sufficiently high agreement
rate (about 85\%) between two human annotators.}

The three main factors in this algorithm are (a) accessible text
regions, (b) semantic consistency, and (c) dynamic syntactic
preference.  The algorithm is invoked for each sentence after the
earlier finite-state transduction phases have determined the best
sequence(s) of nominal and verbal expressions.  Crucially, each
nominal expression is associated with a set of template data objects
that record various linguistic and textual attributes of the referring
expressions contained in it. These data objects are similar to {\em
discourse referents} in discourse semantics 
\cite{Karttunen:DiscRef,Kamp:81,Heim:PhD,kamp+reyle:1993}, in that
anaphoric expressions such as {\it she} are also associated with
corresponding anaphoric entities. A pleonastic {\it it} has no
associated entities.  Quantificational nominals such as {\it each
company} are associated with entity objects because they are
`anaphoric' to group entities accessible in the context.  In this
setup, the effect of reference resolution is {\em merging} of
multiple entity objects. Here is the algorithm.

\ben
\item INPUT: Template entities with the following textual, syntactic,
 and semantic features:

\ben
\item determiner type (e.g., DEF, INDEF, PRON)
\item grammatical or numerical number (e.g., SG, PL, 3)
\item head string  (e.g., {\it automaker})
\item the head string sort in a sort hierarchy 
(e.g., {\tt automaker}\ra {\tt company}\ra {\tt organization})
\item modifier strings (e.g., {\it newly founded}, {\it
with the pilots})
\item text span (the start and end byte positions)
\item sentence and paragraph positions\footnote{Higher text
structure properties such as subsections and sections should also be
considered if there are any. Exact accessibility computation using
complex hierarchical text structures is a future topic of study.}
\item text region type (e.g., HEADLINE, TEXT)
\een

\item FOR EACH potentially 
anaphoric entity object in the current sentence, in the
left-to-right order, DO

\ben
\item[(1)] COLLECT antecedent entity objects from the {\em accessible text
region}.

\bit
\item 
For an entity in a HEADLINE text region, the
entire TEXT is accessible 
because the headline summarizes the 
text. 
\item
For an entity in a TEXT region, 
everything {\em preceding} its text span is accessible (except for
the HEADLINE). 
Intrasentential cataphora is allowed only for
first-person pronouns.
\item In addition, a {\em locality} 
assumption on anaphora sets a (soft) window of
search for each referring expression type---the
entire preceding text for {\em proper names}, narrower for {\em
definite noun phrases}, even narrower for {\em pronouns}, and only the
current sentence for {\em reflexives}. 
In the MUC-6 system, the window size was arbitrarily set to ten
sentences
for definites and three sentences for pronouns, ignoring
paragraph boundaries, and no antecedents beyond the limit were considered. This clearly left ample room for refinement.\footnote{For example, in a more recent FASTUS system, paragraphs are
also considered in setting the limit, and 
at most one candidate beyond the limit is proposed 
when no candidates are found
within the limit.}

\eit

\item[(2)] FILTER with {\em semantic consistency} between
the anaphoric entity $E_1$ and the potential antecedent entity $E_2$.

\bit

\item {\em Number Consistency}: 
$E_1$'s number must be consistent with $E_2$'s number---for example, twelve
is consistent with PLURAL, but not with SINGULAR. As a special case,
plural pronouns ({\it they, we}) can take singular organization
antecedents.

\item {\em Sort Consistency}: 
$E_1$'s sort must either EQUAL or SUBSUME $E_2$'s sort. This reflects
a {\em monotonicity} assumption on anaphora---for example, since {\tt
company} subsumes {\tt automaker}, {\it the company} can take {\it a
Chicago-based automaker} as an antecedent, but it is too risky to
allow {\it the automaker} to take {\it a Chicago-based company} as an
antecedent.  On the other hand, since {\tt automaker} and {\tt
airline} are neither the same sort nor in a subsumption relation, {\it
an automaker} and {\it the airline} cannot corefer. (The system's sort
hierarchy is still sparse and incomplete.)

\item {\em Modifier Consistency}: 
$E_1$'s modifiers must be consistent with $E_2$'s modifiers---for example,
{\it French} and {\it British} are inconsistent, but {\it French} and
{\it multinational} are consistent. (The system doesn't have enough
knowledge to do this well.)  

\eit

\item[(3)] ORDER by {\em dynamic syntactic preference}.  
The following ordering
approximates the relative {\em salience} of entities.  The basic
underlying hypothesis is that intrasentential candidates are more
salient than intersentential candidates as proposed, for example, in
\namecite{hobbs:1978} and \namecite{kameyama:intra}, and that
fine-grained syntax-based salience fades with time.  Since
fine-grained syntax with grammatical functions is unavailable, the
syntactic prominence of subjects and left-dislocation is approximated
by the left-right linear ordering.

\ben
\item the preceding part of the {\em same sentence} in the {\bf
left-right} order
\item the {\em immediately preceding sentence} in the {\bf left-right} order
\item other preceding sentences within the `limit' (see above) in the
{\bf right-left} order
\een
\een

\item OUTPUT: 
After each anaphoric entity has found 
an ordered set of potential antecedent
entities, there are destructive (indefeasible) and 
nondestructive (defeasible) options.
\ben
\item {\em Destructive Option}: MERGE the anaphoric entity into the preferred antecedent entity .
\item {\em Nondestructive Option}: RECORD the antecedent entity list in
the anaphoric entity to allow reordering (i.e., {\em preference revisions}) by 
event merging or overall model selection. 
\een
The MUC-6 system took the destructive option. 
The nondestructive option has been implemented in a more recent system.
\een

These basic steps of ``COLLECT, FILTER, and ORDER by
salience'' are analogous to Lappin and Leass's (1994) pronoun
resolution algorithm, but each step in FASTUS 
relies on considerably poorer
syntactic input.  The present algorithm thus provides an
interesting case of what happens with extremely poor syntactic input,
even poorer than in Kennedy and Boguraev's (1996) system. This
comparison will be discussed later.

\subsection{Name Alias Recognition}

In addition to the above general algorithm, a special-purpose {\em
alias recognition} algorithm is invoked for coreference resolution of proper names.\footnote{In addition, a specific-type name may be converted 
into another type in certain linguistic contexts.
For instance, in
{\it a subsidiary of Mrs. Field},
{\it Mrs. Field} is converted from a person name into a company name.} 

\ben
\item INPUT: The input English text is in mixed cases. An earlier
transduction phase has recognized {\tt unknown
names} as well as specific-type names for {\tt
persons}, {\tt locations}, or {\tt organizations} using
name-internal pattern matching and known name lists.

\item FOR EACH new sentence, FOR EACH {\tt unknown name},
IF it is an alias or acronym of another name already recognized in the
given text, MERGE the two---an {\em alias} is a selective substring of
the full name (e.g., {\it Colonial} for {\it Colonial Beef}), and {\em
acronym} is a selective sequence of initial characters in the full
name (e.g., {\it GM} for {\it General Motors}).

\een

\section{Overall Performance}

The MUC-6 FASTUS reference resolution algorithm handled only
coreference (i.e., identity of reference) of proper names, definites,
and pronouns. These are the `core' anaphoric expression types whose
dependencies tend to be constrained by surface textual factors such as
locality. The MUC-6 Coreference Task evaluation included coreference
of bare nominals, possessed nominals, and indefinites as well, which
the system did not handle because we didn't have a reliable algorithm
for these mostly `accidental' coreferences that seemed to require
deeper inferences.  Nevertheless, the system scored a recall of 59\%
and precision of 72\% in the blind evaluation of thirty newspaper
articles.

Table~\ref{core} shows the system's performance in resolving the core
discourse anaphors in five randomely selected articles from the
development set. Only five articles were examined here because the
process was highly time-consuming. The performance for each expression
type varies widely from article to article because of unexpected
features of the articles. For instance, one of the five articles is a
letter to the editor with a text structure drastically
different from news reports. On average, 
we see that the resolution
accuracy (i.e., recall) was the highest for proper names (69\%),
followed by pronouns (62\%) and definites (46\%).
There were not enough instances of reflexives to compare.

\begin{table}
\begin{small}
\begin{center}
\begin{tabular}{||l|r|r||} \hline \hline
Expression Type	& Number of 	& Correctly\\
		& Occurrences 	& Resolved\\
\hline
Definites & 61     	& 28(46\%)\\
Pronouns  & 39      	& 24(62\%)\\
Proper Names & 32    	& 22(69\%)\\
Reflexives & 1		& 1(100\%)\\
\hline
TOTAL & 133 	& 75(56\%)\\
\hline \hline
\end{tabular}
\caption{Core Discourse Anaphors in Five Articles}\label{core}
\end{center}
\end{small}
\end{table}

Table~\ref{pronoun} shows the system's performance for pronouns
broken down by two parameters, grammatical person and
inter- vs. intrasentential antecedent. The system did quite well
(78\%) with third-person pronouns with intrasentential antecedents,
the largest class of such pronouns.

\begin{table}
\begin{small}
\begin{center}
\begin{tabular}{||l|l|r|r||} \hline \hline
Grammatical	& Intra/Inter-S 	& Number of & Correctly\\
Person		& Antecedent	& Occurrences & Resolved\\
\hline
3rd person & intra-S   & 27     & 21(78\%)\\
3rd person & inter-S   & 6	& 2(33\%)\\
{\it that} & inter-S   & 1	& 0(0\%)\\
1st/2nd person & inter-S   & 5  & 1(20\%)\\
\hline
reflexive  & intra-S   & 1      & 1(100\%)\\
\hline \hline
\end{tabular}
\caption{Pronouns in Five Articles}\label{pronoun}
\end{center}
\end{small}
\end{table}

Part of the pronoun resolution performance here enables a preliminary
comparison with the results reported in (1)
\namecite{lappin+leass:1994} and (2) \namecite{kennedy+boguraev:1996}.
For the third-person pronouns and reflexives, the performance was (1)
86\% of 560 cases in five computer manuals and (2) 75\% of 306 cases
in twenty-seven Web page texts. The present FASTUS system correctly
resolved 71\% of 34 cases in five newspaper articles. This progressive
decline in performance corresponds to the progressive decline in the
amount of syntactic information in the input to reference
resolution. To summarize the latter decline,
\namecite{lappin+leass:1994} 
had the following components in their algorithm.
\ben
\item INPUT: fully parsed sentences with grammatical roles and 
head-argument and head-adjunct relations
\item Intrasentential syntactic filter based on syntactic noncoreference
\item Morphological filter based on person, number, and gender features
\item Pleonastic pronoun recognition
\item Intrasentential binding for reflexives and reciprocals
\item Salience computation based on grammatical role, grammatical parallelism, frequency of mention, proximity, and sentence recency
\item Global salience computation for noun phrases (NPs) in equivalence classes (with seven salience factors)
\item Decision procedure for choosing among equally preferred candidate antecedents
\een
\namecite{kennedy+boguraev:1996} 
approximated the above components with a poorer syntactic input, which
is an output of a part-of-speech tagger with grammatical function
information, plus NPs recognized by finite-state patterns and NPs'
adjunct and subordination contexts recognized by heuristics. With this
input, grammatical functions and precedence relations were used to
approximate 2 and 5. Finite-state patterns approximated 4. Three
additional salience factors were used in 7, and a preference for
intraclausal antecedents was added in 6; 3 and 8 were the same.

The present algorithm works with an even poorer syntactic input, as
summarized here.  
\ben
\item INPUT: a set of finite-state approximations of sentence parts, which can be overlapping or discontiguous, with
no grammatical function, subordination, or adjunct
information. 
\item No disjoint reference filter is used.
\item Morphological filter is used.
\item Pleonastic pronouns are recognized with finite-state patterns. 
\item Reflexives
simply limit the search to the current sentence, with no attempt at
recognizing coarguments. No reciprocals are treated.
\item Salience is approximated by computation based on linear order and recency. No grammatical parallelism is recognized.
\item Equivalence classes correspond to {\em merged entity objects} whose `current' positions are always the most recent mentions.
\item Candidates are deterministically ordered, so no decision procedure is needed.
\een
Given how {\it little} syntactic information is used in FASTUS
reference resolution, the 71\% accuracy in pronoun resolution is
perhaps unexpectedly high.  This perhaps shows that 
linear ordering and recency are major indicators of salience, especially
because grammatical functions correspond to constituent ordering in
English.  The lack of disjoint reference filter is not the most
frequent source of errors, and a coarse-grained treatment of
reflexives does not hurt very much, mainly because of the infrequency of
reflexives.

\section{An Example Analysis}

In the IE context, the task of entity reference resolution is to
recognize referential links among partially described entities within
and across documents, which goes beyond third-person pronouns and
identity of reference.  The expression types to be handled include
bare nominals, possessed nominals, and indefinites, whose referential
links tend to be more `accidental' than textually signaled, and the
referential links to be recognized include subset, membership, and
part-whole.

Consider one of the five articles evaluated, the one with the most
number and variety of referential links, 
for which FASTUS's performance was the
poorest. Even for the `core'
anaphoric expressions of 20 definites, 14 pronouns, and 7 names
limited to coreference, for which the code was prepared, the recall
was only 51\%.  

Figure~\ref{example} shows this article annotated with
referential indices.  The same index indicates coreference.  Index
subscripts (such as $4a$) indicate {\em subset, part}, or {\em
membership} of another expression (e.g., indexed $4$).  The index
number ordering, 1,...,N, has no significance.  Each sentence in the
TEXT region is numbered with paragraph and sentence numbers, so, for
instance, 2-1 is the first sentence in the second paragraph.

\begin{small}
\begin{figure*}
\ben
\item[] HEADLINE: {\bf American Airlines}$_1$ Calls for {\bf Mediation}$_{15}$
In {\bf Its}$_1$ {\bf Union}$_4$ {\bf Talks}$_2$

\item[] DOCUMENT DATE: 02/09/87

\item[] SOURCE: WALL STREET JOURNAL

\item[1-1] Amr corp.'s {\bf American Airlines}$_1$ unit said {\bf
it}$_1$ has called for {\bf federal
mediation}$_{15}$ in {\bf its}$_1$ {\bf contract talks}$_2$ with {\bf
unions}$_4$ representing {\bf its}$_1$ {\bf pilots}$_{16}$
and {\bf flight attendants}$_{17}$. 
\item[2-1]
A spokesman for {\bf the company}$_1$ said {\bf American}$_1$
officials ``felt {\bf talks}$_2$
had reached a point where {\bf mediation}$_{15}$ would be helpful.'' 
\item[2-2] {\bf Negotiations}$_{2a}$ with {\bf the pilots}$_{16}$ have been going on for 11 months;
{\bf talks}$_{2b}$ with {\bf flight attendants}$_{17}$ began six months ago.
\item[3-1]
{\bf The president}$_5$ of {\bf the Association of Professional Flight
Attendants}$_{4a}$, which represents {\bf American}$_1$'s {\bf more than 10,000 flight
attendants}$_{17}$, called {\bf the request}$_3$ for {\bf mediation}$_{15}$ ``premature'' and
characterized {\bf it}$_3$ as a bargaining tactic that could lead to a
lockout. 
\item[3-2] {\bf Patt gibbs}$_5$, {\bf president}$_5$ of {\bf the
association}$_{4a}$, said {\bf talks}$_{2b}$ with {\bf the
company}$_1$  seemed to be progressing well and {\bf the call}$_3$ for
{\bf mediation}$_{15}$ came
as a surprise. 
\item[4-1]
The major outstanding issue in {\bf the negotiations}$_{2b}$ with {\bf the flight
attendants}$_{17}$ is a two-tier wage scale, in which recent employees'
salaries increase on a different scale than the salaries of employees
who have worked at {\bf american}$_1$ for a longer time.
\item[4-2] {\bf The union}$_{4a}$  wants to narrow the differences between the new scale
and the old one.
\item[5-1]
{\bf The company}$_1$ declined to comment on {\bf the negotiations}$_{2b}$ or the
outstanding issues.
\item[5-2]
Representatives for {\bf the 5,400-member Allied Pilots Association}$_{4b}$
didn't return phone calls. 
\item[6-1]
Under the Federal Railway Labor Act, 
[if {\bf the mediator}$_{15a}$ fails to
bring {\bf the two sides}$_7$ together and {\bf the two sides}$_7$  don't agree to binding
arbitration, a 30-day cooling-off period follows]$_6$. 
\item[6-2]
After {\bf that}$_6$, {\bf the union}$_{7a}$ can strike or {\bf the company}$_{7a}$ can lock {\bf the union}$_{7a}$ 
out. 
\item[7-1]
{\bf Ms. Gibbs}$_5$ said that in response to {\bf the company}$_1$'s
move, {\bf her}$_5$ {\bf union}$_{4a}$
will be ``escalating'' {\bf its}$_{4a}$ ``corporate campaign'' against {\bf American}$_1$ over
the next couple of months. 
\item[7-2]
In {\bf a corporate campaign}$_{10}$, {\bf a union}$_9$ tries to get
{\bf a company}$_8$'s
financiers, investors, directors and other financial partners to
pressure {\bf the company}$_8$ to meet {\bf union}$_9$ demands. 
\item[8-1]
{\bf A corporate campaign}$_{10}$, {\bf she}$_5$ said, appeals to {\bf
her}$_5$ {\bf members}$_{17}$ because ``{\bf it}$_{10}$
is a nice, clean way to take a job action, and {\bf our}$_{4a}$ {\bf women}$_{17}$ are hired to
be nice.'' 
\item[8-2] {\bf the union}$_{4a}$ has decided not to strike, {\bf she}$_5$ said. 
\item[9-1]
{\bf The union}$_{4a}$ has hired {\bf a number of professional
consultants}$_{14}$ in {\bf its}$_{4a}$
{\bf battle}$_{18}$ with {\bf the company}$_1$, including {\bf Ray Rogers}$_{14a}$ of Corporate Campaign
Inc., {\bf the New York labor consultant}$_{14a}$ who developed {\bf the strategy}$_{12}$ at
{\bf Geo. A. Hormel \& Co.}$_{13}$'s Austin, Minn., meatpacking plant last
year. 
\item[9-2]
{\bf That campaign}$_{12}$, which included a strike, faltered when {\bf the company}$_{13}$
hired new workers and the International Meatpacking Union wrested
control of the local union from {\bf Rogers}$_{14a}$' supporters. 
\een
\caption{\label{example}Example Article Annotated with Referential Links}
\end{figure*}
\end{small}

Note that not all of these referential links need be
recognized for each particular IE application. However, since
reference resolution must consider all of the text for any particular
application for the reason mentioned above, it is reasonable to assume
that an ideal domain-independent reference resolution component should
be able to recognize all of these. 
Is this a realistic goal, especially in an IE context? 
This question is left open for now.



\subsection{Error Analysis}

Table~\ref{reflinks} shows the system's performance in recognizing the
referential links in this article, grouped by referring expression
types. These exclude the initial mention of each referential chain.
Notable sources of errors and necessary extensions are summarized for each expression type here.

\begin{table}
\begin{small}
\begin{center}
\begin{tabular}{||l|r|r||} \hline \hline
Expression Type	& Number of 	& Correctly\\
		& Occurrences 	& Resolved\\
\hline
Definites & 25      	& 10(40\%)\\
Pronouns  & 14      	& 10(71\%)\\
Bare Nominals  & 12 	& 3(25\%)\\
Proper Names & 7    	& 2(29\%)\\
Possessed Nominals & 6 	& 0(0\%)\\
Indefinites & 3 	& 0(0\%)\\
\hline
TOTAL & 67 	& 25(37\%)\\
\hline \hline
\end{tabular}
\caption{Referential Links in the Example}\label{reflinks}
\end{center}
\end{small}
\end{table}

\bdes
\item[Pronouns:] Of the four pronoun resolution errors, one is due to a parse error ({\it American} in 7-1 was
incorrectly parsed as a person entity, to which {\it she} in 8-1 was
resolved), {\it that} in 6-2 is a {\it discourse deixis} 
\cite{webber:1988}, beyond the scope of the current approach, 
and two errors 
({\it it} in 3-1 and {\it its} in 7-1) 
were due to the left-right ordering of intrasentential candidates.  
Recognition of parallelism among clause conjuncts and a stricter locality preference for possessive pronouns may help here.

\item[Definites:]
Of the fifteen incorrect resolutions of definites,
five have {\em nonidentity} referential
relationships, and hence were not handled. 
These nonidentity cases must be handled to avoid erroneous
identity-only resolutions. 
Two errors were due to the failure to distinguish between generic and
specific events. Token-referring definites ({\it the union} in 8-2 and
{\it the company} in 9-1) were incorrectly resolved to recently
mentioned types.
Three errors were due to the failure to recognize {\em synonyms}
between, for example, {\it call} (3-2) vs. {\it request} (3-1) and {\it campaign} (9-1) vs. {\it strategy} (9-1).
Other error sources are a failure in recognizing an appositive
pattern (9-1), the left-right ordering of the candidates in the 
previous sentence (9-2), and three `bugs'.

\item[Proper Names:]
Name alias recognition was unusually poor (2 out of
7) because {\it American} was parsed as a person-denoting noun. The
lowercase print in {\it Patt gibbs} also made it difficult to link
it with {\it Ms. Gibbs}. Such {\em parse errors} and {\em input
anomalies} hurt performance.

\item[Bare Nominals:] Since bare nominals were not explicitly resolved, the only correct resolutions (3 out of 12) were due to recognition of appositive patterns. How can the other cases be treated?
We need to understand the discourse semantics of bare
nominals better before developing an effective algorithm.

\item[Possessed Nominals:]
A systematic treatment of possessed nominals is a necessary
extension. The basic algorithm will look like this---resolve the
possessor entity $A$ of the possessed nominal $A's\ B$, then if there
is already an entity of type $B$ {\em associated with} $A$ mentioned
{\it anywhere in the preceding text}, then this is the referent of
$B$. Possessed nominal resolution also 
requires a `synonymy' computation to resolve,
for example, {\it its battle} (9-1) to {\it its corporate campaign} (7-1).
It also needs `inferences' that rely on multiple successful resolutions.
For instance, {\it her members} in 8-1 must first resolve
{\it her} to {\it Ms. Gibbs}, who is {\it president of the
association}, and this `association' is {\it the Association of
Professional Flight Attendants}. After it is understood as `the
members of the Association of Professional Flight Attendants,' the
coreference with {\it the flight attendants} can be
inferred. Similarly for {\it our women} in 8-1.

\item[Indefinites:] Some indefinites are `coreferential' to generic types, for example, {\it a corporate campaign} (7-2, 8-1).
Given the difficulty in distinguishing between generic and
specific event descriptions, it is unclear whether it will ever be
treated in a systematic way.

\edes

\section{Conclusions}

In an operational end-to-end discourse understanding system, a
reference resolution component must work with input data containing
parse errors, lexicon gaps, and mistakes made by earlier reference
resolution. In a state-of-the-art IE system such as SRI's FASTUS,
reference resolution must work with considerably impoverished
syntactic analysis of the input sentences.
The present reference resolution approach within an IE system is
robust and efficient, and performs pronoun resolution to an almost
comparable level with a high-accuracy algorithm in the literature.
Desirable extensions include nonidentity referential
relationships, treatments of bare nominals, possessed nominals, and
indefinites, type-token distinctions, and recognition of synonyms.
Another future direction is to turn this component into a corpus-based
statistical approach using the relevant factors identified in the
rule-based approach.  
The need for a large tagged corpus may be
difficult to satisfy, however.

\section{Acknowledgment}
This work was in part supported by U.S. government contracts.

\begin{small}

\begin{thebibliography}{}

\bibitem[\protect\citename{Appelt \bgroup et al.\egroup }1995]{fastus4}
Appelt, Douglas, Jerry Hobbs, John Bear, David Israel, Megumi Kameyama, Andy
  Kehler, David Martin, Karen Myers, and Mabry Tyson.
\newblock 1995.
\newblock {SRI International} {FASTUS} system: {MUC-6} test results and
  analysis.
\newblock In {\em Proceedings of the 6th Message Understanding Conference},
  pages 237--248. DARPA.

\bibitem[\protect\citename{Grosz, Joshi, and Weinstein}1995]{GJW:1995}
Grosz, Barbara, Aravind Joshi, and Scott Weinstein.
\newblock 1995.
\newblock Centering: A framework for modelling the local coherence of
  discourse.
\newblock {\em Computational Linguistics}, 21(2):203--226.

\bibitem[\protect\citename{Heim}1982]{Heim:PhD}
Heim, Irene.
\newblock 1982.
\newblock {\em The Semantics of Definite and Indefinite Noun Phrases}.
\newblock {Ph.D.} thesis, University of Massachusetts at Amherst.

\bibitem[\protect\citename{Hobbs}1978]{hobbs:1978}
Hobbs, Jerry.
\newblock 1978.
\newblock Resolving pronoun references.
\newblock {\em Lingua}, 44:311--338.
\newblock Also in B.~Grosz, K.~Sparck-Jones, and B.~Webber, eds., {\it Readings
  in Natural Language Processing}, Morgan Kaufmann, Los Altos, CA, 1986,
  339-352.

\bibitem[\protect\citename{Hobbs \bgroup et al.\egroup }1996]{fastus}
Hobbs, Jerry~R., Douglas~E. Appelt, John Bear, David Israel, Megumi Kameyama,
  Mark Stickel, and Mabry Tyson.
\newblock 1996.
\newblock {FASTUS}: A cascaded finite-state transducer for extracting
  information from natural-language text.
\newblock In E.~Roche and Y.~Schabes, editors, {\em Finite State Devices for
  Natural Language Processing}. MIT Press, Cambridge,Massachusetts.

\bibitem[\protect\citename{Kameyama}in press]{kameyama:intra}
Kameyama, Megumi.
\newblock in press.
\newblock Intrasentential centering: A case study.
\newblock In Marilyn Walker, Aravind Joshi, and Ellen Prince, editors, {\em
  Centering in Discourse}. Oxford University Press.

\bibitem[\protect\citename{Kamp}1981]{Kamp:81}
Kamp, Hans.
\newblock 1981.
\newblock A theory of truth and semantic representation.
\newblock In J.~Groenendijk, T.~Janssen, and M.~Stokhof, editors, {\em Formal
  Methods in the Study of Language}. Mathematical Center, Amsterdam, pages
  277--322.

\bibitem[\protect\citename{Kamp and Reyle}1993]{kamp+reyle:1993}
Kamp, Hans and Uwe Reyle.
\newblock 1993.
\newblock {\em From Discourse to Logic}.
\newblock Kluwer, Dordrecht.

\bibitem[\protect\citename{Karttunen}1976]{Karttunen:DiscRef}
Karttunen, Lauri.
\newblock 1976.
\newblock Discourse referents.
\newblock In James~D. McCawley, editor, {\em Syntax and Semantics: {N}otes from
  the Linguistic Underground}, volume~7. Academic Press, New York, pages
  363--386.

\bibitem[\protect\citename{Kennedy and Boguraev}1996]{kennedy+boguraev:1996}
Kennedy, Christopher and Branimir Boguraev.
\newblock 1996.
\newblock Anaphora for everyone: Pronominal anaphora resolution without a
  parser.
\newblock In {\em Proceedings of the 16th International Conference on
  Computational Linguistics (COLING-'96)}. Association for Computational
  Linguistics.

\bibitem[\protect\citename{Lappin and Leass}1994]{lappin+leass:1994}
Lappin, Shalom and Herbert Leass.
\newblock 1994.
\newblock An algorithm for pronominal anaphora resolution.
\newblock {\em Computational Linguistics}, 20(4):535--562.

\bibitem[\protect\citename{Sidner}1979]{Sidner:Towards}
Sidner, Candace.
\newblock 1979.
\newblock Towards a computational theory of definite anaphora comprehension in
  {E}nglish discourse.
\newblock Technical Report 537, MIT Artificial Intelligence Laboratory,
  Cambridge, MA, June.

\bibitem[\protect\citename{Sundheim}1995]{beth-muc6}
Sundheim, Beth.
\newblock 1995.
\newblock Overview of results of the {MUC}-6 evaluation.
\newblock In {\em Proceedings of the 6th Message Understanding Conference},
  pages 13--32. DARPA.

\bibitem[\protect\citename{Webber}1988]{webber:1988}
Webber, Bonnie.
\newblock 1988.
\newblock Discourse deixis: Reference to discourse segments.
\newblock In {\em Proceedings of the 26th Annual Meeting of the Association for
  Computational Linguistics}, pages 113--122. Association for Computational
  Linguistics, June.

\bibitem[\protect\citename{Webber}1978]{Webber:PhD}
Webber, Bonnie~Lynn.
\newblock 1978.
\newblock {\em A Formal Approach to Discourse Anaphora}.
\newblock {Ph.D.} thesis, Harvard University.

\end{thebibliography}

\end{small}

\end{document}